\documentclass[10pt,twocolumn,letterpaper]{article}

\usepackage[pagenumbers]{cvpr} 

\usepackage{graphicx}
\usepackage{amsmath}
\usepackage{amssymb}
\usepackage{booktabs}
\usepackage[accsupp]{axessibility}  
\newcommand{\fref}[1]{Figure~\ref{#1}}

\newcommand{\tref}[1]{Table~\ref{#1}}

\DeclareMathOperator{\relu}{ReLU}

\usepackage[pagebackref,breaklinks,colorlinks]{hyperref}

\usepackage[capitalize]{cleveref}
\crefname{section}{Sec.}{Secs.}
\Crefname{section}{Section}{Sections}
\Crefname{table}{Table}{Tables}
\crefname{table}{Tab.}{Tabs.}

\def\cvprPaperID{6798} 
\def\confName{CVPR}
\def\confYear{2022}

\begin{document}

\title{AirObject: A Temporally Evolving Graph Embedding for Object Identification}

\author{Nikhil Varma Keetha$^{1, 2}$ \and Chen Wang$^{1}$ \and Yuheng Qiu$^{1}$ \and Kuan Xu$^{3}$ \and Sebastian Scherer$^{1}$ \and
$^{1}$Carnegie Mellon University \hspace{6mm} $^{2}$IIT (ISM) Dhanbad, India \hspace{6mm} $^{3}$Geek+ Corp\\
{\tt\small \{keethanikhil, xukuanhit\}@gmail.com, chenwang@dr.com, \{yuhengq, basti\}@andrew.cmu.edu}
}
\maketitle

\begin{abstract}

Object encoding and identification are vital for robotic tasks such as autonomous exploration, semantic scene understanding, and re-localization. Previous approaches have attempted to either track objects or generate descriptors for object identification. However, such systems are limited to a ``fixed" partial object representation from a single viewpoint. In a robot exploration setup, there is a requirement for a temporally ``evolving” global object representation built as the robot observes the object from multiple viewpoints. Furthermore, given the vast distribution of unknown novel objects in the real world, the object identification process must be class-agnostic. In this context, we propose a novel temporal $3$D object encoding approach, dubbed AirObject, to obtain global keypoint graph-based embeddings of objects. Specifically, the global $3$D object embeddings are generated using a temporal convolutional network across structural information of multiple frames obtained from a graph attention-based encoding method.
We demonstrate that AirObject achieves the state-of-the-art performance for video object identification and is robust to severe occlusion, perceptual aliasing, viewpoint shift, deformation, and scale transform, outperforming the state-of-the-art single-frame and sequential descriptors.
To the best of our knowledge, AirObject is one of the first temporal object encoding methods.
Source code is available at \url{https://github.com/Nik-V9/AirObject}.

\end{abstract}

\section{Introduction}

\begin{figure}
\centering
\begin{tabular}{cccc}
\includegraphics[width=1\linewidth]{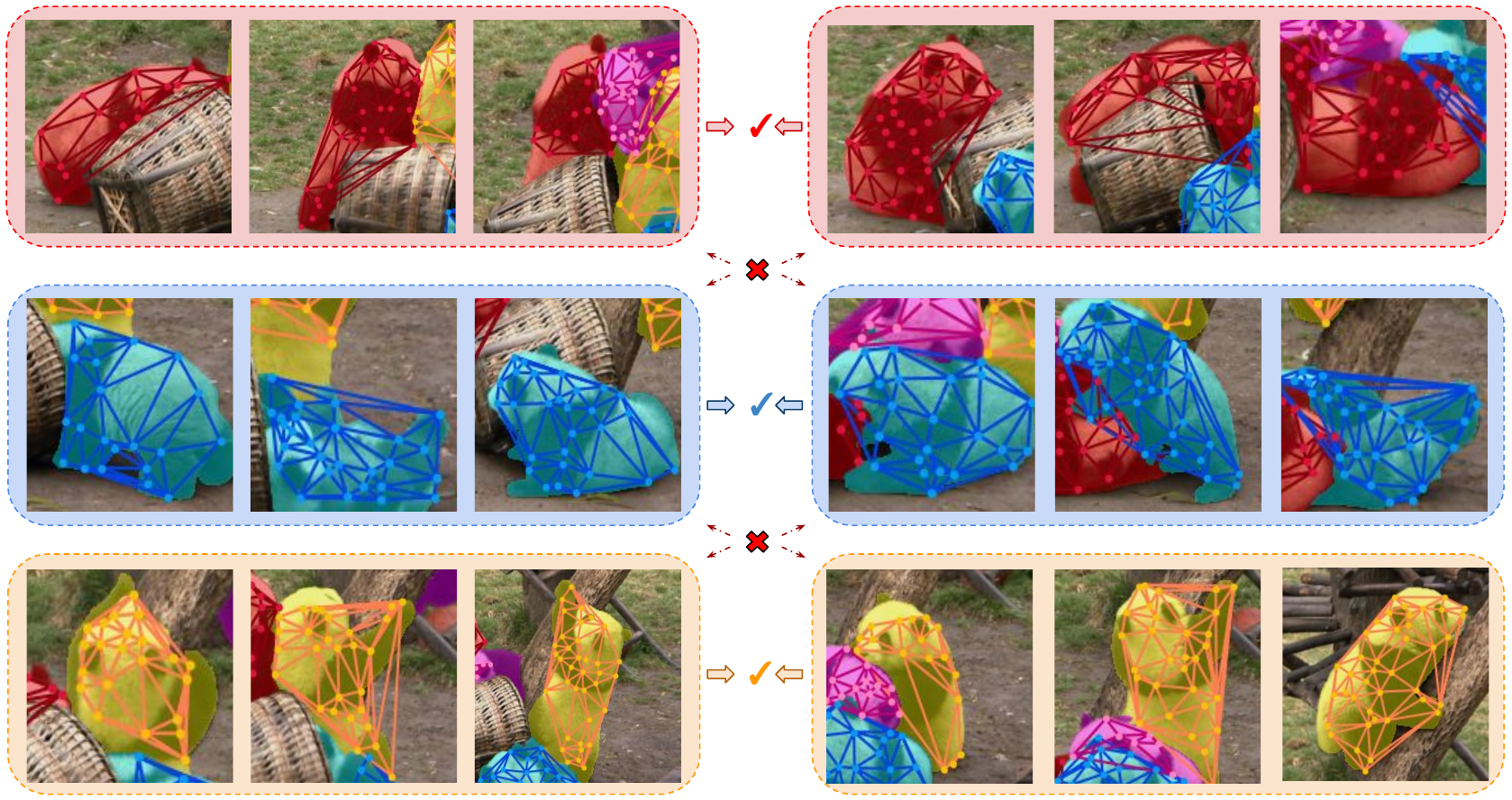}
\end{tabular}
\caption{Temporally evolving topological graph representations of objects within a video sequence. We propose a method, AirObject, to match these temporally evolving representations and alleviate problems caused by perceptually-aliased occluded single frame representations.}
\label{fig:Overview}
\end{figure}

Object encoding and identification are crucial for robotic tasks such as autonomous exploration, semantic scene understanding, and loop closure in simultaneous localization and mapping (SLAM). For example, object-based semantic SLAM and identification of revisited objects require robust and efficient object encodings~\cite{sharma2021compositional,wang2020visual,wang2021unsupervised}. Prior approaches proposed in the literature have attempted to track object detections~\cite{wang2007simultaneous}, use keypoint features~\cite{dubuisson1994modified}, and generate graph-based embeddings for object matching~\cite{xu2022aircode}. However, such systems are limited to a ``fixed" object representation from a single viewpoint and are not robust to severe occlusion, viewpoint shift, perceptual aliasing, or scale transform. These single frame representations tend to lead to false correspondences amongst perceptually-aliased objects, especially when severely occluded. Hence, a robust object encoding method that aggregates the temporally ``evolving" object structures is necessary since we often observe more information when the camera or object moves, as shown in \fref{fig:Overview}.

In this work, we propose a novel temporal encoding approach, dubbed AirObject, that encapsulates the evolving topological graph-based representations of objects.
It is very simple and only contains three modules.
Specifically, we use extracted deep learned keypoint features~\cite{detone2018superpoint} across multiple frames to form sequences of object-wise topological graph neural networks (GNNs), which on embedding generate temporal $3$D object descriptors. We next employ a graph attention-based sparse encoding method on these topological GNNs to generate content graph features and location graph features representing the structural information of the object. Then, these graph features are aggregated across multiple frames using a single-layer temporal convolutional network to generate a temporal $3$D object descriptor. These generated object descriptors, which accumulate knowledge across multiple evolving representations of the objects, are robust to severe occlusion, viewpoint changes, deformation, perceptual aliasing, and the scale transform.

In summary, we make the following contributions:

\begin{itemize}
    \item To learn the geometric relationship of the keypoints, we construct topological object graphs for each frame using Delaunay triangulation.
    \item We introduce a simple yet effective temporal object encoding method to aggregate and embed multiple instances into an object descriptor.
    \item Extensive experiments show that AirObject consistently provides the state-of-the-art performance for video object identification on four large-scale datasets.
\end{itemize}

\section{Related Work}

In this section, we review single-frame and sequential methods based on handcrafted and deep learned features. Furthermore, we also review Visual Place Recognition (VPR) methods that can be extended to object identification. Object tracking methods based on networks such as Mask R-CNN~\cite{he2017mask} are not included because they are unsuitable for object re-identification.

\subsection{Single-frame Representations}

Handcrafted features such as SIFT~\cite{lowe2004distinctive} and SURF~\cite{bay2008speeded} have been widely used in classical approaches for loop closure, object matching, and VPR. One such classical approach, fast appearance-based mapping (FABMAP)~\cite{glover2012openfabmap}, utilizes the trained visual vocabulary of SURF features, obtained by hierarchical k-means clustering, for identifying revisited objects through feature distribution matching. Further extending this idea, a binary descriptor ORB~\cite{rublee2011orb} was used in DBoW2~\cite{galvez2012bags} to achieve better speed. Further building on the notion of vocabulary-based retrieval, several approaches~\cite{garcia2018ibow, schlegel2018hbst, gehrig2017visual, leutenegger2011brisk} have been proposed. However, these handcrafted features based methods are sensitive to environmental changes and lead to false matches when the local descriptors are not discriminative enough.

\vspace{0.5em}

The recent success of convolutional neural networks (CNN)~\cite{krizhevsky2012imagenet} in computer vision has led to the rise of deep learned features-based image retrieval. The methods employing deep learned features have shown tremendous improvements over handcrafted features. One such method, \cite{chen2017deep}, uses a multi-scale feature encoding across two CNN architectures to generate CNN features that are viewpoint invariant, thereby providing considerable performance improvement. Another popular end-to-end deep learning-based approach, NetVLAD~\cite{arandjelovic2016netvlad} generates descriptors inspired by the traditional vector of locally aggregated descriptors (VLAD). Further exploring other input modalities such as RGB-D images and point cloud data, several approaches~\cite{schwarz2015rgb, zaki2016convolutional, zaki2019viewpoint} have attempted to incorporate spatial/depth data into the RBG domain for object recognition.

Recently proposed deep learning method, SuperPoint~\cite{detone2018superpoint} leverages a self-supervised framework to train interest point detectors and descriptor extractors. Further building on SuperPoint, SuperGlue~\cite{sarlin2020superglue} introduced a local feature matcher based on graph attention~\cite{velivckovic2017graph} where the interest points are nodes of a graph, and their associated descriptors are the node features. Both SuperPoint and SuperGlue have been widely adopted for the task of feature matching and hierarchical VPR~\cite{sarlin2019coarse, keetha2021hierarchical}. Similar to SuperGlue, Xu~\etal embeds object-wise fully connected graph-based representations of SuperPoint features using a Sparse Object Encoder to generate object descriptors~\cite{xu2022aircode}. However, this approach doesn't consider the explicit geometry available from the SuperPoint interest points. It is limited to only a single viewpoint, making it susceptible to false matches under perceptual aliasing and occlusion. In this context, our proposed framework generates temporal object descriptors that aggregate structural knowledge across multiple object instances.

\subsection{Sequential Representations}

While single frame representations have been extensively used across the literature, temporal information for compact representations has received limited attention in robotics, especially for loop closure. However, there exists an extensive array of Spatio-temporal representation techniques in related fields of research~\cite{jalal2017robust, girdhar2017actionvlad, wu20193, dai2017shape, yang2020spatial}. Many of these approaches use LSTM~\cite{hochreiter1997long}, GRUs~\cite{chung2014empirical}, Graph Convolutional Networks (GCNs)~\cite{kipf2016semi}, and Temporal Convolutional Networks~\cite{bai2018empirical} to model the temporal relations. In the context of VPR, there have been a few approaches attempting to leverage Spatio-temporal information in terms of landmarks~\cite{johns2011place} and bio-inspired memory cells~\cite{nguyen2013spatio}.

Recently, Facil~\etal presented concatenation, fusion, and recurrence for learning sequential representations~\cite{facil2019condition}. Before this work, the concatenation of descriptors within a sequence was explored~\cite{arroyo2015towards}, where binarization was used for efficient place recognition. Similarly, Neubert~\etal employs recurrence for learning a bio-inspired topological map of the environment~\cite{neubert2019neurologically}. More recently, a temporal convolutional network, SeqNet, was proposed to learn sequential descriptors from single image descriptors for hierarchical VPR~\cite{garg2021seqnet}. However, the learned temporal information for SeqNet depends on the underlying visual attributes from single image descriptors which don't provide explicit knowledge regarding structural relations. In this context, our approach learns temporal information across evolving topological graph representations, which provide spatial/structural information regarding the object.

\begin{figure}
\centering
\begin{tabular}{cc}
\includegraphics[width=0.45\textwidth]{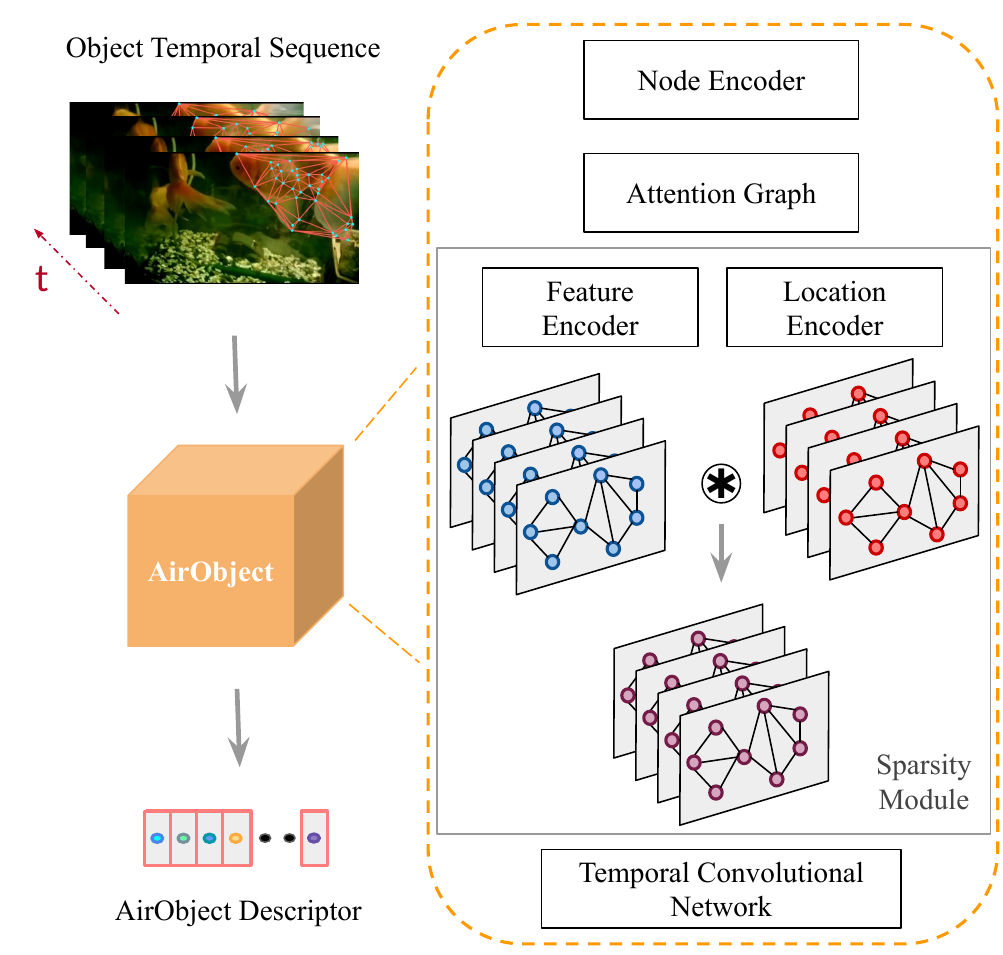} \\
\end{tabular}
\caption{Schematic of our proposed approach. The SuperPoint based temporal topological graph representations of an object are input to the graph attention encoder to generate content graph features and location graph features. Then, these features are node-wise multiplied and temporally aggregated to generate the AirObject descriptor.}
\label{fig:Approach}
\end{figure}

\section{Proposed Approach}

We propose a new architecture, AirObject, to encode objects from a video, which is shown in \fref{fig:Approach}.
In this section, we will first present the topological graph representations of objects, and then describe the structure of the graph attention encoder and a temporal encoding approach to generate the AirObject descriptors. Finally, we discuss the loss functions to supervise the encoders.

\subsection{Topological Graph Representations}

Intuitively, a group of feature points on an object form a graphical representation where the feature points are nodes and their associated descriptors are the node features. Essentially, the graph's nodes are the distinctive local features of the object, while the edges/structure of the graph represents the global structure of the object. We believe that embedding such a topological graph-based representation of an object containing both distinctive local features and global object structure will enable robust object identification similar to humans~\cite{tarr2017concurrent}. Hence based on this hypothesis, we formulate a procedure to generate topological graph representations of feature points corresponding to objects.

Given an object, we extract a set of feature points corresponding to the object, where the position of each feature point is denoted as $p_i = (x,y), i \in [1,N]$ and the associated descriptor as $d_i \in \mathbb{R}^{D_p}$, where $D_p$ is the descriptor dimension. In practice, these object-wise grouped feature points could be obtained using point detector SuperPoint~\cite{detone2018superpoint} along with ground-truth instance segmentations or masks from an off-the-shelf network like Mask R-CNN~\cite{he2017mask} or an open-world object detector~\cite{joseph2021towards}. Given these object-wise grouped feature points, our goal is to generate a topological graph representation that leverages the explicit geometry provided by the feature point positions.

\begin{figure}
\centering
\begin{tabular}{cc}
\includegraphics[width=0.22\textwidth]{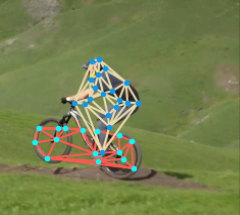}
&
\includegraphics[width=0.22\textwidth]{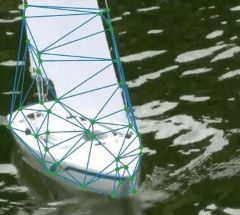}
\end{tabular}
\caption{Topological Graph Representations of Objects. These representations are generated by using Delaunay Triangulation on object-wise grouped SuperPoint keypoints.}
\label{fig:triangulation}
\end{figure}

We build a topological graph structure for the feature points corresponding to an object by using Delaunay triangulation~\cite{lee1980two} on the positions of the feature points as shown in \fref{fig:triangulation}. Delaunay triangulation is a mathematical formulation, where given a set of discrete points, the objective is to provide a triangulation avoiding narrow and intersecting triangles such that no discrete point is present inside the circumcircle of any triangle of the triangulation. This particular property coupled with fast compute time makes it suited to generate a triangular mesh representation that intuitively captures the feature points' local \& global structure. We believe that this mesh representation will enable the graph attention encoder to reason better about the object's structure, thereby making the final temporal object descriptor robust to deformation or occlusion.

\subsection{Graph Attention Encoder}

The topological graph representations of the object feature points are inputted to the Graph Attention-based sparse object encoder~\cite{xu2022aircode}. A particular desirable property of a sparse object encoder is that a single keypoint should only have a local effect on sparse locations of the object descriptor so that addition or removal of keypoints doesn't significantly change the object descriptor. Furthermore, the object encoder should encode distinct keypoints to unique locations within the object descriptor. To facilitate this, the encoder comprises a Node Encoder, a two-layer Graph Attention (GAT)~\cite{velivckovic2017graph} module followed by a Sparsity module that contains two parallel heads, namely a feature and a location encoder whose outputs are element-wise multiplied.

The Node Encoder takes the positions and descriptors of the topological graph-based object representation and encodes the graph's node features $x_i$ as the concatenation of the point descriptor and a transform of the point position:
\begin{equation}
\centering
    {x_i}^{(1)} = [d_i \parallel \mathrm{MLP}(p_i)], \ \ {x_i}^{(1)} \in \mathbb{R}^{D_n},
\end{equation}
where $D_n = D_p + D_m$, $\parallel$ denotes the concatenation operator, the Multi-layer Perceptron (MLP) module maps $\mathbb{R}^{2} \mapsto \mathbb{R}^{D_m}$, and ${x_i}^{l}$ denotes the output of the $l$-th layer of the GNN. In practice, we normalize the position $p_i$ to $[-1,1]$ by the object size, which considers the object center as the origin. The concatenation of the transformation of the position as opposed to a summation similar to SuperGlue~\cite{sarlin2020superglue} helps the encoder explicitly learn the object structure since the relative position information is not blended into the descriptor. Furthermore, this concatenation operation enables the Sparsity module of the encoder to learn sparse non-zero locations for the keypoints based on the object structure.

The graph's node features $x_i$ and the adjacency matrix from the topological graph structure are input to a two-layered GAT to enable structured attention-based message propagation between the features of the object\footnote{Refer to \cite{velivckovic2017graph} for more details on GNNs and GAT.}. This helps the encoder reason about global feature interactions between the distinctive local keypoint-based features of the object. Afterward, the output of the GAT is passed on to the Sparsity module to encode the graph embeddings such that learned location features of the node decide the sparse location of the keypoint on the temporal object descriptor.

The Sparsity module of the encoder contains two parallel heads where each head has two stacked sparsity layers to learn the location feature ${x_i}^{L}$ and the content feature ${x_i}^{C}$, whose inputs are $x_i$ from the GAT (we leave out the layer index $(l)$ for simplicity). The sparsity layers for the location node features and content node features are defined as:

\begin{equation}
\centering
    _{}^{(l+1)}{x_i}^{L} = \relu({W_L}^{(l)}._{}^{(l)}{x_i}^{L}),
\tag{2a}
\end{equation}
\begin{equation}
\centering
    _{}^{(l+1)}{x_i}^{C} = \relu({W_C}^{(l)}._{}^{(l)}{x_i}^{C}),
\tag{2b}
\end{equation}
where ${W_L}^{(l)}, \ {W_C}^{(l)} \in \mathbb{R}^{D_o \times {D_n}^{l}}, \ {D_n}^{l} < D_o$ are the learnable location and content weights, respectively, and $D_o$ is the dimension of the temporal object descriptor. Then, these location and content features are node-wise multiplied to generate the structural graph features ${x_i}^{S}$.

\subsection{Temporal Encoding}

Given a sequence of topological graph object representations input to the graph attention encoder, we first obtain a sequence of structural graph features. Then, these structural graph features are node-wise stacked to form a tensor of size ($N_s \times D_o$) where $N_s$ is the number of structural graph feature nodes across a sequence, and $D_o$ is the temporal object descriptor dimension. To perform temporal aggregation and encode these features into a single object descriptor, we use a Temporal Convolutional Network (TCN) comprising of a single layer 1-D convolution (with bias, without padding) followed by a Sequence Average Pooling (SAP) layer and an L2-Normalization layer.

In the convolution layer, we use a $1$-D filter of length $1$ that operates with stride $1$ across the $N_s$ dimension of the input tensor. We use a filter length of $1$ to ensure that the encoding process is not limited to a fixed sequence length and is compatible with the varying total number of structural graph nodes. The dimension of the structural graph features, $D_o$, forms the input channels (feature maps) for the convolution layer while the output channels are also set to $D_o$. Thus, the convolutional kernel is of size $D_o \times 1 \times D_o$. The output of the $1$-D convolution layer is further input to an SAP layer across the $N_s$ dimension to obtain a descriptor of size $1 \times D_o$. To ensure compatibility with cosine similarity, we then use an L2-Normalization across the $D_o$ dimension to obtain the final temporal AirObject descriptor $A_k$.

\subsection{Loss Functions}

The Graph Attention encoder is supervised by a sparse location loss, dense feature loss~\cite{xu2022aircode}. The objective of the sparse location loss is to ensure that location feature ${x_i}^{L}$ is a sparse vector. The sparse location loss $L_s$ is defined as the $l_1$-norm of ${x_i}^{L}$.
\begin{equation}
\centering
    {L_s} = \sum_{i=1}^{N} \lVert \phi ({x_i}^{L}) {\rVert}_1,
\tag{3}
\end{equation}
where $\phi (x) = x / \lVert x {\rVert}_2$ is a $l_2$-normalization to prevent the location features from being zero. Given that the sparse location loss ensures that keypoints are encoded into sparse locations on the object descriptor, the objective of the dense feature loss is to ensure that distinctive keypoints are encoded to unique sparse locations on the object descriptor. Hence, dense feature loss $L_d$ is defined as the negative $l_1$-norm of the location features.
\begin{equation}
\centering
    {L_d} = \max \left( 0, \delta - \phi \left( \lVert \sum_{i=1}^{N}  ({x_i}^{L}) {\rVert}_1 \right) \right),
\tag{4}
\end{equation}
where $\delta > 0$ is a positive constant. Intuitively, the combined optimization of both sparse location loss and dense feature loss enables the object encoder to encode graph representations such that the similar keypoints are encoded to similar locations, while distinctive keypoints cover different locations retaining the density of the object descriptor. 

\begin{table*}
\centering
\caption{Quantitative Results: Performance comparison on four datasets.}
\scalebox{0.8}{
\begin{tabular}{lcccccccccccc}
\toprule
& \multicolumn{3}{c}{\textbf{YT-VIS}}      & \multicolumn{3}{c}{\textbf{UVO}}       & \multicolumn{3}{c}{\textbf{OVIS}}         & \multicolumn{3}{c}{\textbf{TAO-VOS}} \\
 \cmidrule(lr{0.5em}){2-4} \cmidrule(lr{0.5em}){5-7} \cmidrule(lr{0.5em}){8-10} \cmidrule(lr{0.5em}){11-13}

\multicolumn{1}{c}{\textbf{Methods}} & Precision & Recall & F$1$ & Precision & Recall & F$1$ & Precision & Recall & F$1$ & Precision & Recall & F$1$ \\

\cmidrule(lr{0.5em}){1-1} \cmidrule(lr{0.5em}){2-4} \cmidrule(lr{0.5em}){5-7}
\cmidrule(lr{0.5em}){8-10} \cmidrule(lr{0.5em}){11-13}

\textbf{Single Frame Descriptors:} \\

$2$D Baseline & 68.93 & 80.93 & 74.45
            & 73.18 & 81.85 & 77.27
            & 29.45 & 68.32 & 41.15
            & 46.80 & 72.49 & 56.87 \\
NetVLAD     & 77.52 & 52.41 & 62.54  
            & 86.19 & 72.15 & 78.55
            & 49.41 & 29.45 & 36.90
            & 78.44 & 35.57 & 48.95 \\
SeqNet ($s_l$ = $1$)   & 71.59 & 66.38 & 68.89    
            & 69.44 & 81.03 & 74.79
            & 39.80 & 44.87 & 42.18
            & 66.94 & 49.79 & 57.10 \\
\textit{Ours}: AirObject ($s_l$ = $1$)
& 79.47    & 73.49    & \textbf{76.36}
& 82.99    & 82.51    & \textbf{82.75}
& 38.58    & 55.40    & \textbf{45.49}
& 66.46    & 60.96    & \textbf{63.59} \\

\cmidrule(lr{0.5em}){1-1} \cmidrule(lr{0.5em}){2-4} \cmidrule(lr{0.5em}){5-7}
\cmidrule(lr{0.5em}){8-10} \cmidrule(lr{0.5em}){11-13}

\textbf{Sequential Descriptors:} \\
$3$D Baseline & 65.14 & 86.48 & 74.31    
              & 73.78 & 77.00 & 75.35
              & 25.58 & 80.29 & 38.80
              & 40.60 & 75.84 & 52.89 \\
SeqNet      &  76.73 & 86.48 & 81.31   
            &  96.57 & 72.63 & 82.91
            &  70.19 & 42.09 & 52.62
            &  71.10 & 71.54 & 71.32 \\
\textit{Ours}: AirObject 
& 85.09  &  82.36  & \textbf{83.70}
& 94.31  &  83.79  & \textbf{88.74}
& 69.86  &  42.47  & \textbf{52.82}
& 72.81  &  71.21  & \textbf{72.00} \\
\bottomrule
\end{tabular}
}
\label{tab:mainResult}
\end{table*}

Finally, the Graph Attention Encoder and Temporal Encoder are supervised for object identification using a triplet style matching loss. The objective of the matching loss $L_m$ is to maximize the cosine similarity of positive object pairs and minimize the cosine similarity of negative object pairs.
\begin{equation}
\centering
\begin{aligned}
{L_m} = & \sum_{\{p,q\} \in P^{+}}^{} (1 - C(A_p, A_q)) \\
      &  + \sum_{\{p,q\} \in P^{-}}^{} \max(0, C(A_p, A_q) - \lambda),
\end{aligned}
\tag{5}
\end{equation}
where $\lambda = 0.2$, $C$ is the cosine similarity, and $P^{+}$ and $P^{-}$ are positive and negative object pairs, respectively.

\section{Experimental Results}

\subsection{Datasets}

For video object identification, we require video object sequences where objects are associated across multiple frames. Hence, to train and evaluate our proposed approach, we used four video instance segmentation datasets: YouTube Video Instance Segmentation (YT-VIS)~\cite{yang2019video}, Unidentified Video Objects (UVO)~\cite{wang2021unidentified}, Occluded Video Instance Segmentation (OVIS)~\cite{qi2021occluded}, and Tracking Any Object with Video Object Segmentation (TAO-VOS)~\cite{Voigtlaender21WACV, dave2020tao}. All these datasets contain a large object vocabulary and various challenging scenarios, including perceptually-aliased occluded objects, as described below:

\textit{1) YT-VIS:} This large-scale dataset contains over $3,000$ high-resolution youtube videos annotated with $\approx 5,000$ unique video instances across a $40$-category object label set including common objects such as animals, vehicles, and persons. Furthermore, it contains a large amount of perceptually-aliased objects in various environmental backdrops, making it challenging. 
We split the sequences into a training split of $2,485$ videos and a test split of $500$ videos.

\textit{2) UVO:} This dataset contains class-agnostic video instance segmentations of open-world objects. In particular, all objects present within the Kinetics400~\cite{kay2017kinetics} dataset videos are densely annotated such that there are about 13 objects per video. The open-world class-agnostic nature of the objects coupled with crowded scenes and complex background motions make this a challenging dataset to test the robustness of our proposed object encoding method. We use the Dense-Annotation training and validation sets containing $393$ videos with ground-truth instances for evaluation.

\textit{3) OVIS:} This large-scale dataset was designed with the philosophy of perceiving occluded objects in videos. So for the same, this dataset contains long videos with severely occluded objects annotated with high-quality instance masks across 25 semantic categories. The severe occlusions, long video duration, and crowded scenes make this dataset very challenging for object identification. We use the training set containing $607$ videos for evaluation.

\textit{4) TAO-VOS:} This dataset is a subset of the Tracking Any Object (TAO) dataset~\cite{dave2020tao} with masks for video object segmentation. TAO is a benchmark federated object tracking dataset comprising videos from $7$ datasets captured in diverse environments. In particular, the large object vocabulary ranging from outdoor vehicles to indoor household objects makes this dataset challenging. We use both the training and validation sets containing $626$ videos with ground-truth instances for evaluation.

\subsection{Implementation Details}

The AirObject configurations are $D_p = 256$, $D_m = 16$, and $D_o = 2048$. To test the generalizability, we train only on the YT-VIS train split and evaluate on all four datasets. Firstly, the Graph Attention Encoder coupled with a single-layer perceptron, pretrained on the COCO dataset~\cite{lin2014microsoft}, is finetuned on the YT-VIS training split for single-frame object matching using the sparse location loss, dense feature loss, and matching loss. For finetuning, we employ a batch size of $16$ and a learning rate of $1\mathrm{e}{-4}$ with Adam optimizer~\cite{kingma2014adam}. Then, we freeze the Graph Attention Encoder and train the Temporal Encoder on the YT-VIS training split using the object matching loss. During training, we use a batch size of $16$ containing object sequences of length $s_l \leq 4$ and a learning rate of $1\mathrm{e}{-4}$ with Adam optimizer.

\begin{figure*}
\centering
\begin{tabular}{cccc}
\includegraphics[width=0.225\textwidth]{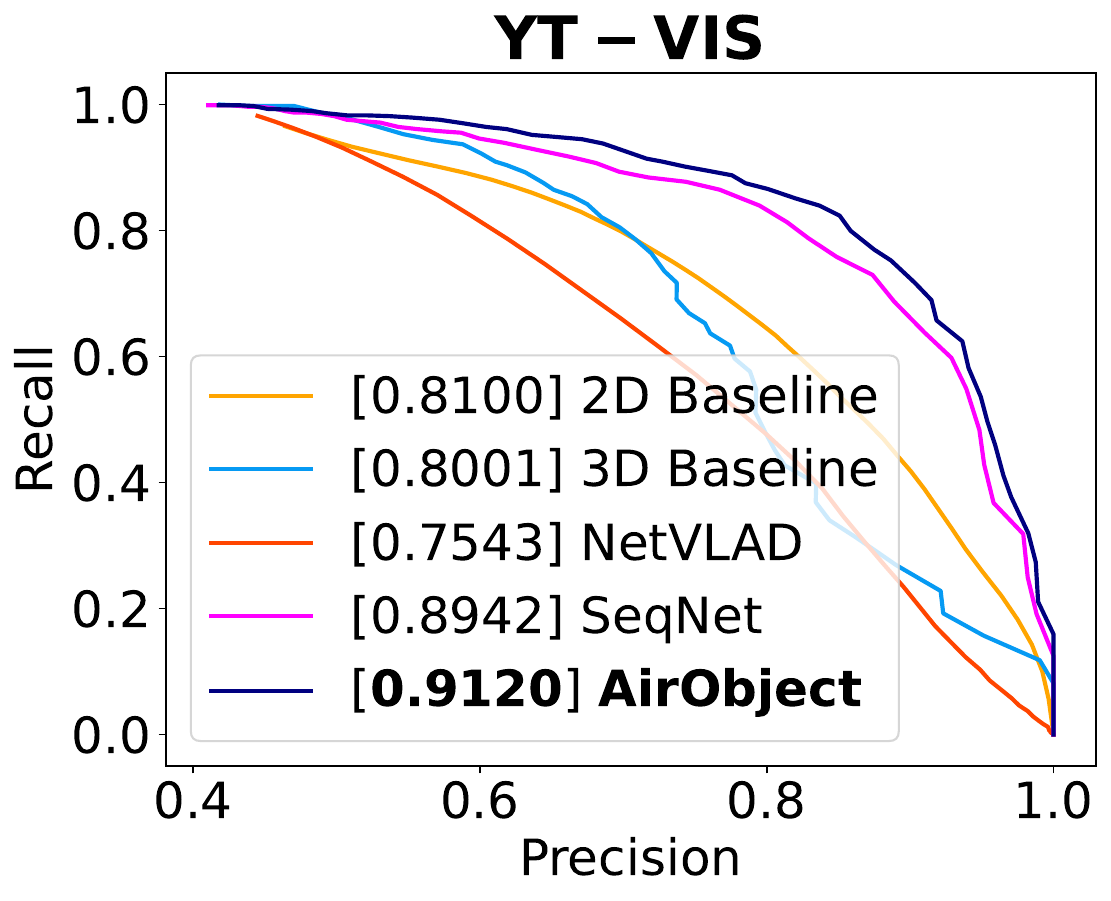}
&
\includegraphics[width=0.225\textwidth]{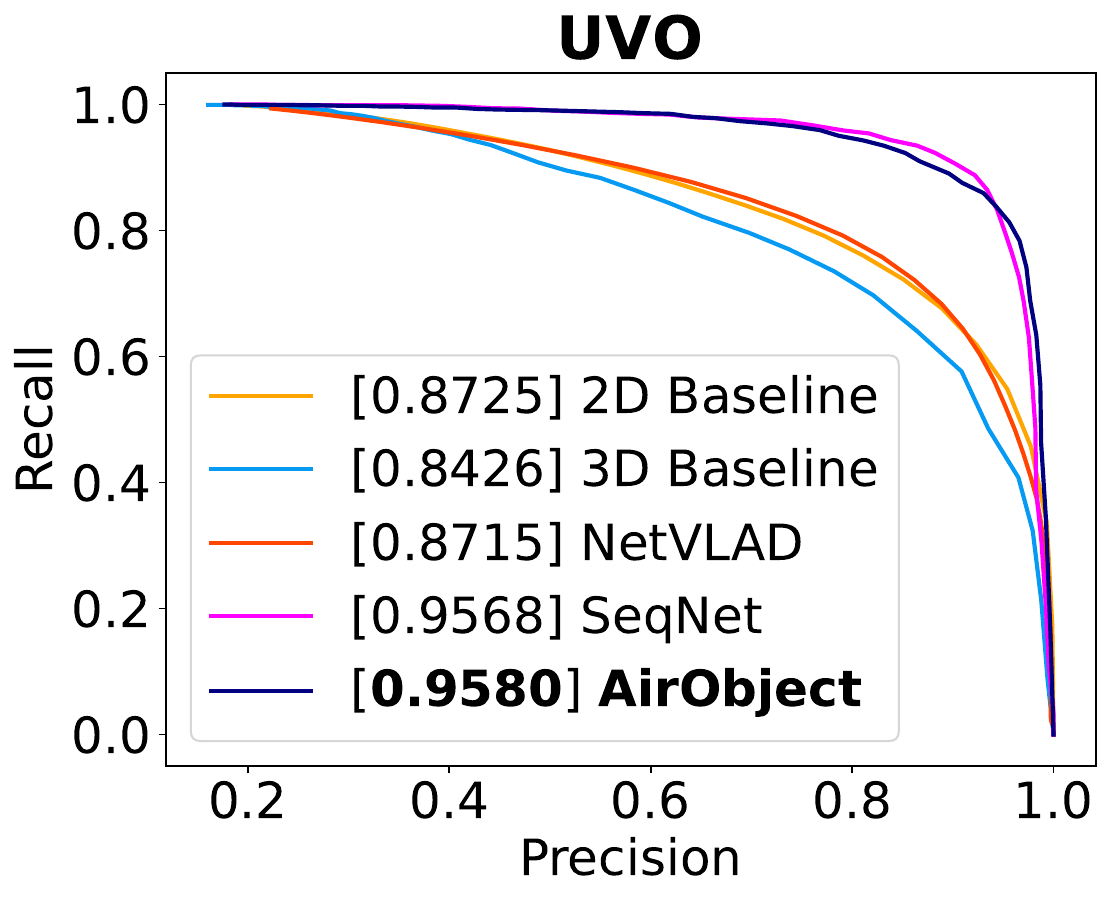}
& 
\includegraphics[width=0.225\textwidth]{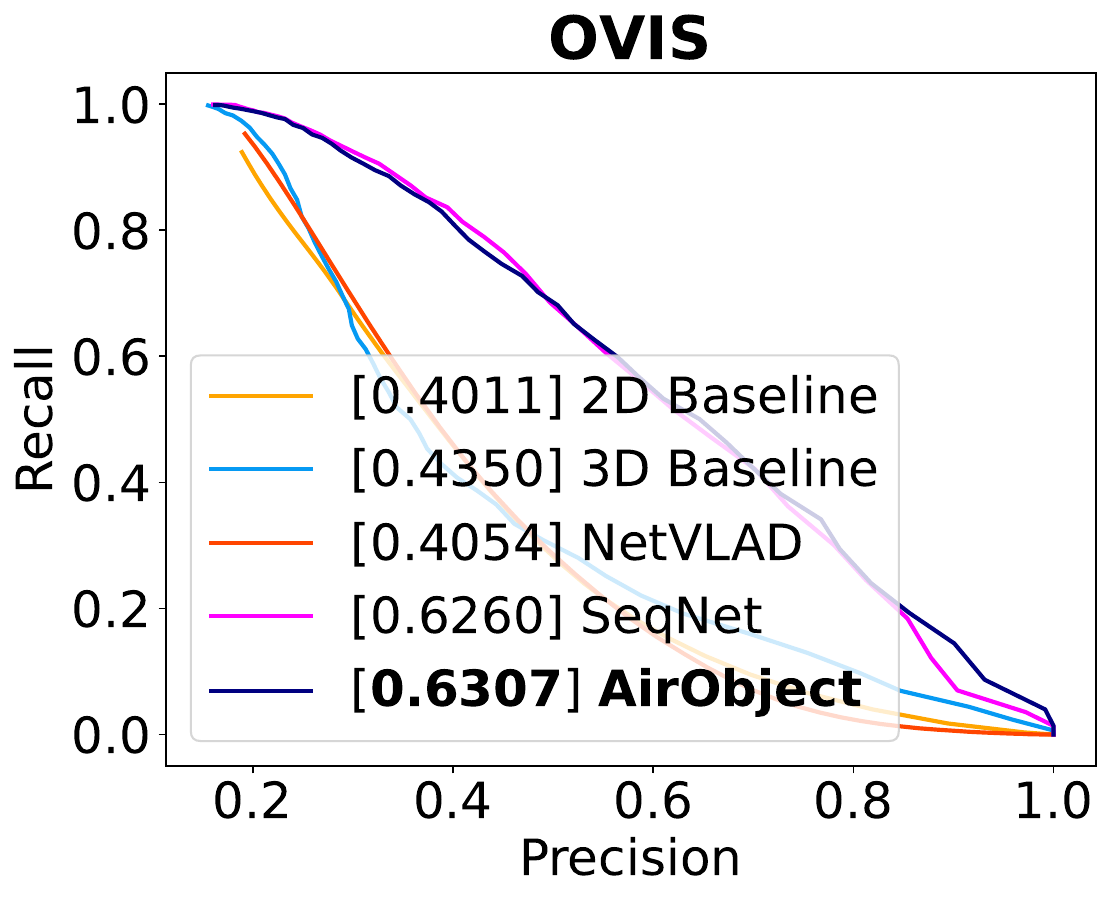}
&
\includegraphics[width=0.225\textwidth]{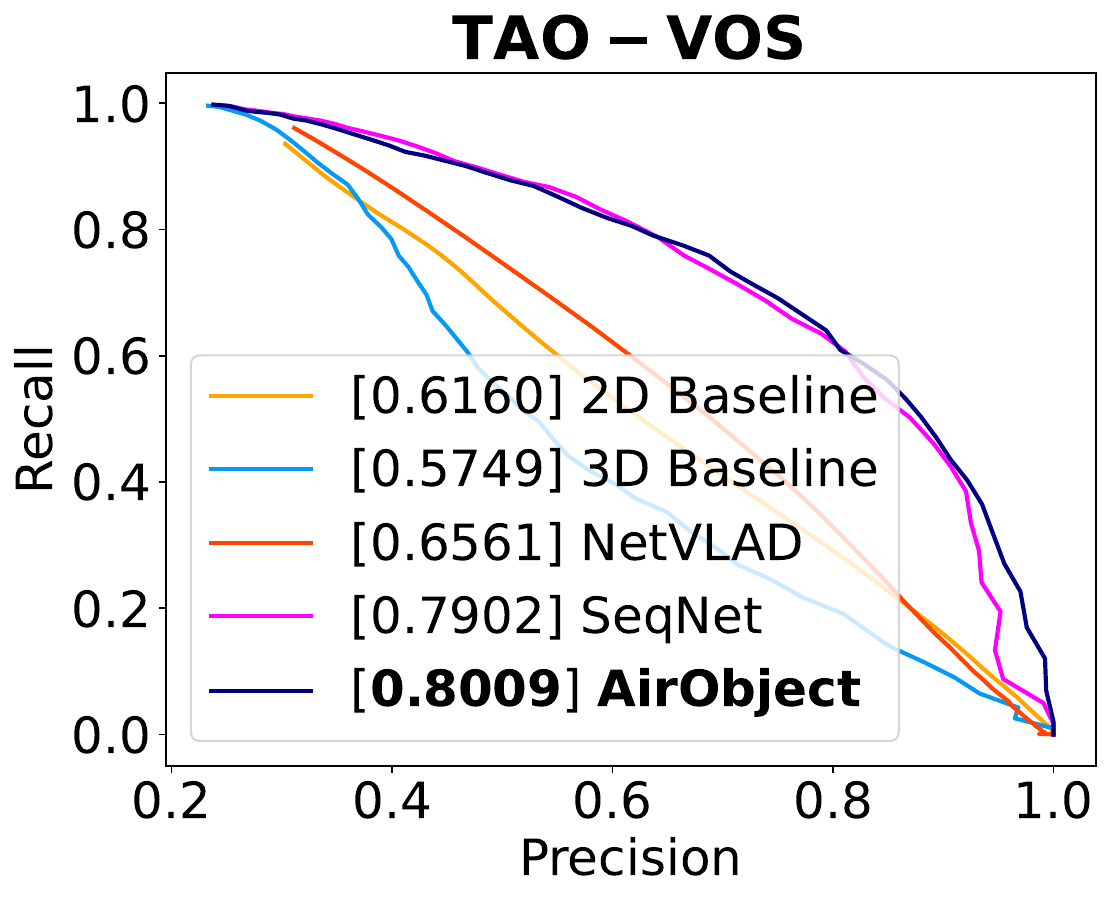} \\
\end{tabular}
\caption{Precision-Recall Curves for Video Object Identification on four datasets. The area under the curve is shown in the [brackets].}
\label{fig:mainResult}
\end{figure*}

\subsection{Evaluation Criteria}

For testing the object identification performance, we consider the first half of the video object sequence as the query and the second half as the reference and evaluate by matching the query and reference object sequences within a video. To determine a match between an object pair, we compute the cosine similarity between the descriptors and define a matching threshold $\rho$. Based on the True Positives and False Positives, we calculate the Precision, Recall, and F1-Score accordingly. Furthermore, by varying the threshold $\rho$ values, we generate Precision-Recall curves and calculate the area under the curves (AUCs).


\begin{figure}
\centering
\begin{tabular}{cc}
\includegraphics[width=0.225\textwidth]{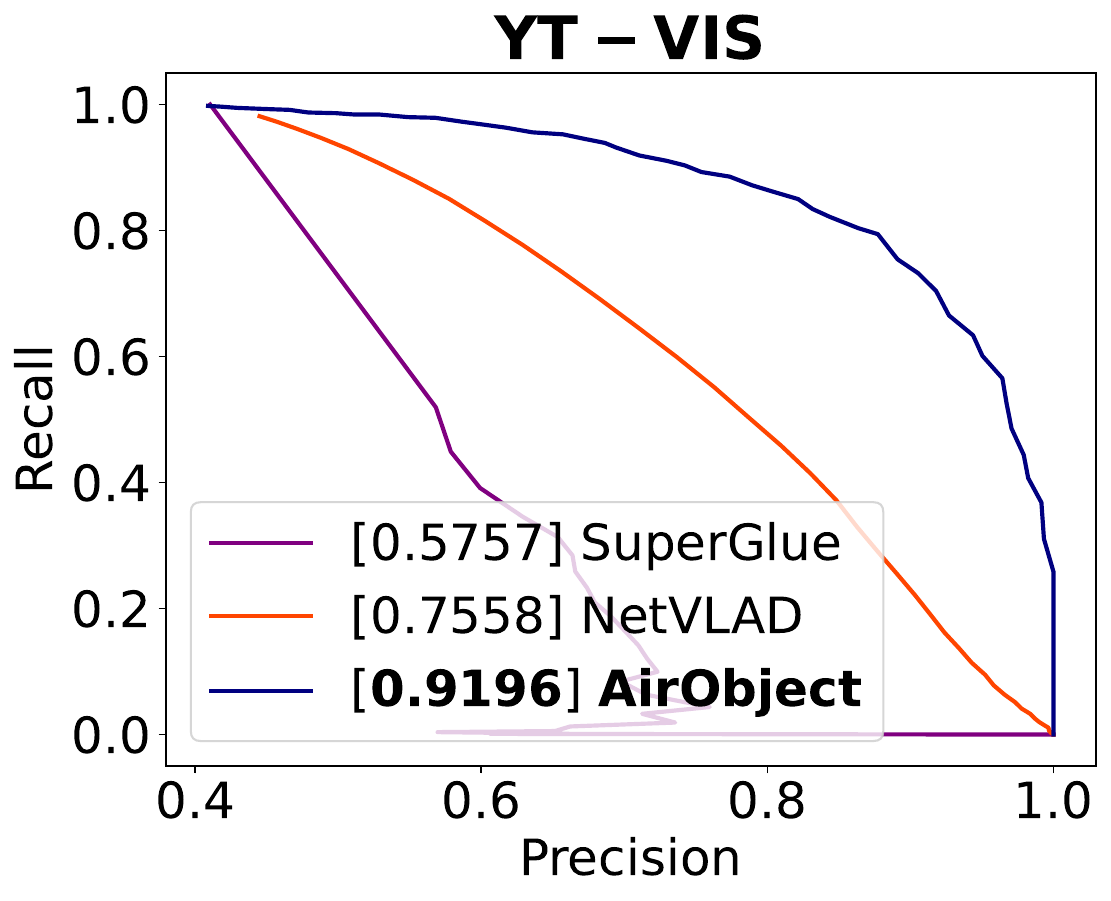}
&
\includegraphics[width=0.225\textwidth]{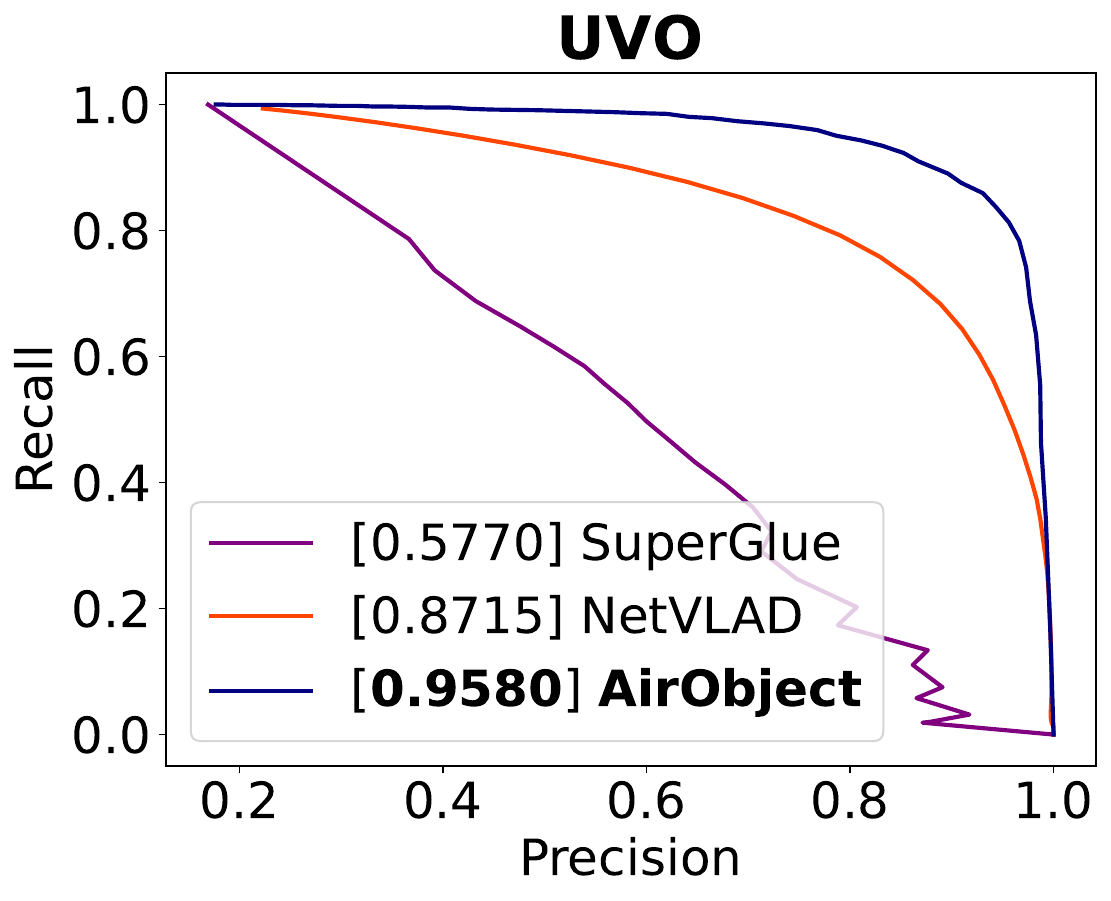}\\
\end{tabular}
\caption{Precision-Recall Comparison with Local Feature Matching (SuperGlue) for Video Object Identification on two datasets. The area under the curve is shown in the [brackets].}
\label{fig:SGResult}
\end{figure}

\subsection{Comparison to State-of-the-art Methods}

We compare our proposed approach against several benchmark baselines: $2$D Baseline, $3$D Baseline, NetVLAD~\cite{arandjelovic2016netvlad}, and SeqNet~\cite{garg2021seqnet}. For the $2$D Baseline, similar to \cite{xu2022aircode}, we use the Graph Attention Encoder coupled with a single layer perceptron to perform single-frame object matching. While, for the $3$D Baseline, we consider the simple case of averaging the $2$D Baseline's single-frame object descriptors to obtain temporal object descriptors.

For the NetVLAD baseline, we obtain object descriptors of size $D_{nv} = 8192$ using NetVLAD ($32$ clusters) on the object-wise grouped features obtained using Superpoint and ground-truth instances. We pre-train NetVLAD on COCO and finetune on the YT-VIS training split. For SeqNet, we use the NetVLAD object descriptors as the backbone and a temporal filter length of $1$ to support varying object sequence lengths. We train SeqNet on the YT-VIS training split with an output descriptor dimension of $D_s = 4096$.

\vspace{0.5em}

\tref{tab:mainResult} and \fref{fig:mainResult} contain quantitative comparisons of AirObject and the baseline methods. It can be observed that AirObject outperforms all the methods both in F-$1$ and AUC consistently across all four datasets. In particular, AirObject outperforms the best performing single-frame and sequential descriptor methods, \ie, $2$D Baseline, NetVLAD, and SeqNet on average by 11.88~\%, 17.58~\%, 2.28~\% for F-$1$ and by 15.05~\%, 15.36~\%, 0.86~\% for AUC respectively. Furthermore, the performance gap between single-frame and temporal methods is consistently large for both metrics, indicating the importance of temporal information. This difference is particularly pronounced for the OVIS and TAO-VOS datasets showing that temporal information encapsulating the evolving object structure helps for mitigating perceptual aliasing and severe occlusions. To further test the applicability of the proposed approach to single-frame object matching, we evaluate SeqNet and AirObject using a sequence length ($s_l$) of 1. From \tref{tab:mainResult}, we can observe that AirObject also provides the best performance for single-frame object identification.

As another baseline, we compare the performance of our method against SuperGlue~\cite{sarlin2020superglue}. While initially proposed for pose estimation and homography, SuperGlue has achieved state-of-the-art performance for feature matching~\cite{keetha2021hierarchical, sarlin2019coarse, hausler2021patch}. So, we extend SuperGlue to object identification using the inlier to outlier ratio as the matching score. From \fref{fig:SGResult}, we can observe that SuperGlue provides high recall and low precision matching. We believe that high perceptual aliasing coupled with no background context leads to low precision for SuperGlue. Furthermore, given the large array of objects present within a single video, the compute time required for local feature matching makes SuperGlue infeasible for object identification. Thus further solidifying the necessity for our proposed robust object encoding and identification method, AirObject.

\begin{figure*}
\centering
\begin{tabular}{cccc}
\includegraphics[width=1\textwidth]{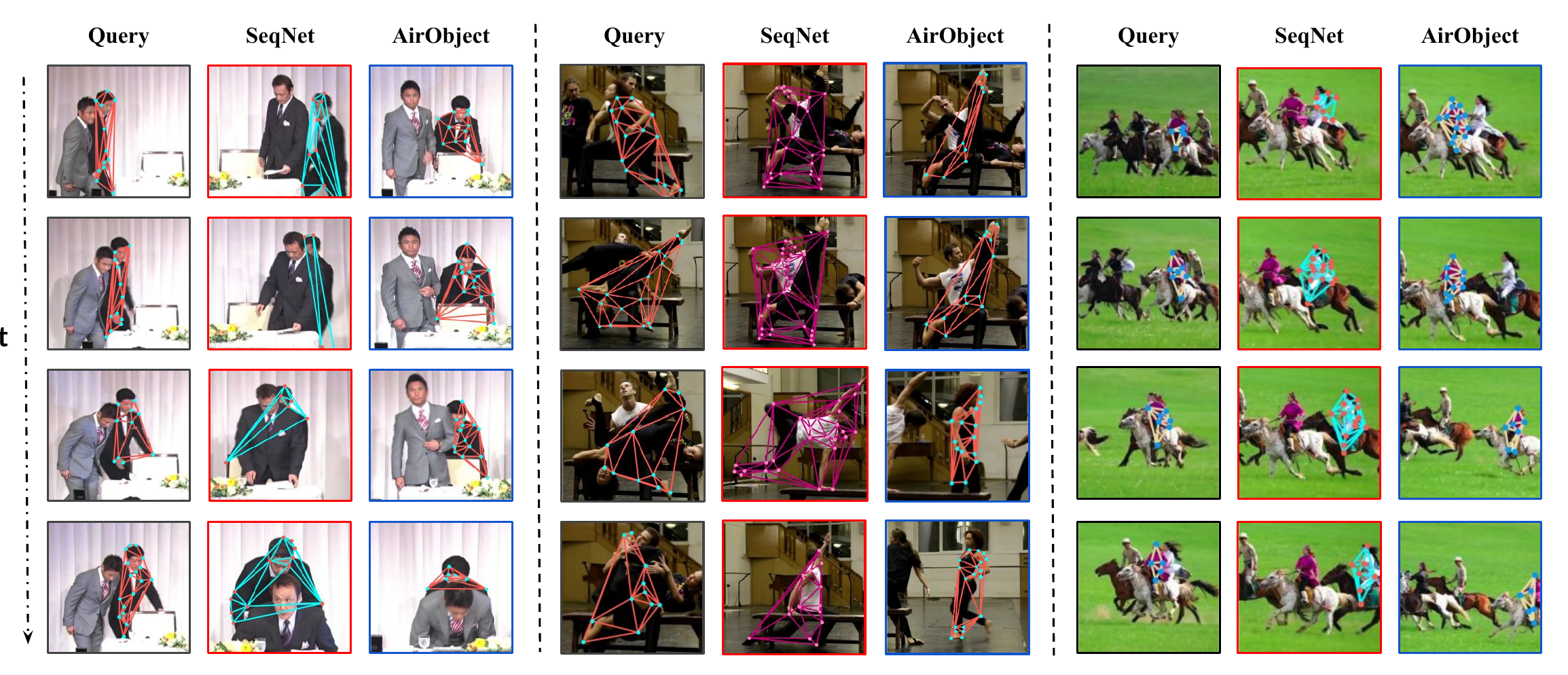} \\
\end{tabular}
\caption{\textbf{Qualitative Results.} In these examples, the proposed AirObject successfully retrieves the matching temporal object sequence, while SeqNet produces incorrect matches. The retrieved objects with our approach are particularly challenging, with a combination of severe occlusion, deformation, perceptual aliasing, and clutter. Please note that SeqNet does not use the graph representations illustrated, and they are only provided for visualization of the object and its structure.}
\label{fig:Qual}
\end{figure*}

In \fref{fig:Qual}, we show a set of example object video sequences, illustrating the matches retrieved with our method compared to SeqNet, along with the topological graph-based object representations.
We believe that SeqNet's dependence on the underlying single-frame descriptor for visual attributes and structural information leads to false matches due to the lack of explicit spatial/structural information. In contrast, AirObject leverages both explicit spatial information from the topological graph representations and temporal information from the object's evolving topology, thereby leading to more robust object identification.

\subsection{Ablation Studies}

\subsubsection{Topological Graph Representations}

\begin{table}
\centering
\caption{Ablation Study: Topological Graph Representations}
\scalebox{0.8}{
\begin{tabular}{lcccc}
\toprule
         & \multicolumn{4}{c}{PR-AUC (\%)} \\
\cmidrule{2-5}
\multicolumn{1}{c}{\textbf{Methods}}  & \textbf{YT-VIS}   & \textbf{UVO}       & \textbf{OVIS}         & \textbf{TAO-VOS} \\
\cmidrule(lr{0.5em}){1-1} \cmidrule(lr{0.5em}){2-2} \cmidrule(lr{0.5em}){3-3} \cmidrule(lr{0.5em}){4-4} \cmidrule(lr{0.5em}){5-5}

2D Baseline (w/o Topological) & 77.87  & 84.96   & 40.82  & 66.03 \\
2D Baseline (w Topological)   & 81.00  & 87.25   & 40.11  & 61.60 \\
AirObject (w/o Topological)   & 86.48  & 93.82   & 63.03  & 79.49 \\
AirObject (w Topological) & \textbf{91.20}    & \textbf{95.80}    & \textbf{63.07}    & \textbf{80.09}  \\
\bottomrule
\end{tabular}
}
\label{tab:topoAblationResult}
\end{table}

To analyze the effectiveness of the topological graph representations within AirObject, we train the $2$D Baseline and AirObject with a fully connected feature points graph as input. From \tref{tab:topoAblationResult}, we can observe that AirObject and $2$D Baseline with topological graph representations outperform those trained with a fully connected graph. This indicates that the topological graph representations help the Graph Attention Encoder reason better about the object geometry, leading to better performance across the four datasets.

\subsubsection{Sequence Length}

\begin{table}
\centering
\caption{Ablation Study: Sequence Length}
\scalebox{0.8}{
\begin{tabular}{lcccc}
\toprule
         & \multicolumn{4}{c}{PR-AUC (\%)} \\
\cmidrule{2-5}
\multicolumn{1}{c}{\textbf{Methods}} & \textbf{YT-VIS}       & \textbf{UVO}       & \textbf{OVIS}         & \textbf{TAO-VOS} \\

\cmidrule(lr{0.5em}){1-1} \cmidrule(lr{0.5em}){2-2} \cmidrule(lr{0.5em}){3-3} \cmidrule(lr{0.5em}){4-4} \cmidrule(lr{0.5em}){5-5}

SeqNet ($s_l$ = $1$)   & 77.65    & 84.10    & 40.47    & 63.57  \\
SeqNet ($s_l \leq 2$)  & 81.21    & 87.95    & 45.50    & 70.48  \\
SeqNet ($s_l \leq 4$)  & 84.22    & 90.58    & 49.69    & 75.26  \\
SeqNet ($s_l \leq 8$)  & 86.46    & 92.53    & 53.42    & 78.35  \\
SeqNet ($s_l \leq 16$) & 88.70    & 93.97    & 56.74    & 79.15  \\
SeqNet                 & 89.42    & 95.68    & 62.60    & 79.02  \\

\cmidrule(lr{0.5em}){1-1} \cmidrule(lr{0.5em}){2-2} \cmidrule(lr{0.5em}){3-3} \cmidrule(lr{0.5em}){4-4} \cmidrule(lr{0.5em}){5-5}

AirObject ($s_l$ = $1$)   & 85.41    & 90.69    & 44.18    & 69.91  \\
AirObject ($s_l \leq 2$)  & 86.56    & 92.13    & 46.40    & 74.16  \\
AirObject ($s_l \leq 4$)  & 88.19    & 93.10    & 48.61    & 77.36  \\
AirObject ($s_l \leq 8$)  & 89.75    & 94.03    & 51.21    & 79.98  \\
AirObject ($s_l \leq 16$) & \textbf{91.39}    & 94.68    & 53.85    & \textbf{81.43}  \\
AirObject  & 91.20    & \textbf{95.80}    & \textbf{63.07} & 80.09  \\
\bottomrule
\end{tabular}
}
\label{tab:seqlenAblationResult}
\end{table}

To verify the effectiveness of temporal information for object identification, we analyze the performance of SeqNet and AirObject for varying object sequence lengths ($s_l$). In line with our intuition, from \tref{tab:seqlenAblationResult}, we can observe that the performance improves as the amount of temporal encoded information increases. Furthermore, the increase in performance tends to decrease  slightly as the sequence length is doubled. Further in line with this trend, for OVIS, the long average video duration (more temporal information) leads to a higher performance gap between AirObject using $s_l \leq 16$ and AirObject using half the object sequence length. Thus, this portrays the efficacy of temporal encoding for object identification.

\subsubsection{Unique Multi-Frame Graph Features}

\begin{table}
\centering
\caption{Ablation Study: Unique Multi-Frame Graph Features}
\scalebox{0.8}{
\begin{tabular}{lcccc}
\toprule
         & \multicolumn{4}{c}{PR-AUC (\%)} \\
\cmidrule{2-5}
\multicolumn{1}{c}{\textbf{Methods}} & \textbf{YT-VIS}       & \textbf{UVO}       & \textbf{OVIS}         & \textbf{TAO-VOS} \\
\cmidrule(lr{0.5em}){1-1} \cmidrule(lr{0.5em}){2-2} \cmidrule(lr{0.5em}){3-3} \cmidrule(lr{0.5em}){4-4} \cmidrule(lr{0.5em}){5-5}

SeqNet   & 89.42    & 95.68    & 62.60    & 79.02  \\
AirObject (Unique Features) & 90.38    & 94.75    & 61.88    & 79.16  \\
AirObject (All Features) & \textbf{91.20}    & \textbf{95.80}    & \textbf{63.07}    & \textbf{80.09}  \\
\bottomrule
\end{tabular}
}
\label{tab:featureAblationResult}
\end{table}

To analyze the effect of recurring object visual features across multiple frames on the temporal descriptor, we perform an experiment encoding only the unique object visual features across the video. In particular, we use the location graph features $x^C$ to identify and encode the unique structural graph features $x^S$ across multiple frames. In \tref{tab:featureAblationResult}, we present results for AirObject using unique multi-frame features and AirObject using all multi-frame features. It can be observed that using unique features across multiple frames leads to a performance drop. We deduce that the recurrence of visual features on an object weighs the temporal object descriptor distribution, making it more unique to that particular object, thereby leading to better performance for AirObject using all multi-frame graph features.

\subsubsection{Temporal Encoder Architecture}

\begin{table}
\centering
\caption{Ablation Study: Temporal Encoder Architecture}
\scalebox{0.725}{
\begin{tabular}{lcccc}
\toprule
         & \multicolumn{4}{c}{PR-AUC (\%)} \\
\cmidrule{2-5}
\multicolumn{1}{c}{\textbf{Methods}} & \textbf{YT-VIS}       & \textbf{UVO}       & \textbf{OVIS}         & \textbf{TAO-VOS} \\
\cmidrule(lr{0.5em}){1-1} \cmidrule(lr{0.5em}){2-2} \cmidrule(lr{0.5em}){3-3} \cmidrule(lr{0.5em}){4-4} \cmidrule(lr{0.5em}){5-5}

AirObject (Single-layer Perceptron) & 84.90    & 92.62    & 51.81    & 68.64  \\
AirObject (GAT) & 85.20    & 88.88    & 49.55    & 70.09  \\
AirObject (Concat. SLP \& SeqNet) & 86.00  & 90.65  & 50.55   & 68.80  \\
AirObject (Attention-based TCN) & 88.77  & 93.38  & 59.22    & 76.61  \\
AirObject (TCN) & \textbf{91.20}    & \textbf{95.80}    & \textbf{63.07}    & \textbf{80.09}  \\
\bottomrule
\end{tabular}
}
\label{tab:encoderAblationResult}
\end{table}

Given the simple nature of our temporal encoder, we further test various model ablations to verify the effectiveness of our design. In particular, we test our TCN, an Attention-based TCN, a Single-layer Perceptron (SLP), and a Graph Attention network (GAT). Inspired by Spatio-temporal work in related fields~\cite{yang2020spatial}, we also test a method for normalized concatenation of descriptors from AirObject (SLP) and SeqNet (with $2$D Baseline backbone). From \tref{tab:encoderAblationResult}, we can observe that the various model ablations, including fully connected attention across multiple temporal states, do not lead to better temporal encoding. Amongst all the methods, using a single-layer TCN works the best, thus proving the simple yet effective nature of our proposed temporal object encoding method, AirObject.

\begin{figure}
\centering
\begin{tabular}{cc}
\includegraphics[width=0.22\textwidth]{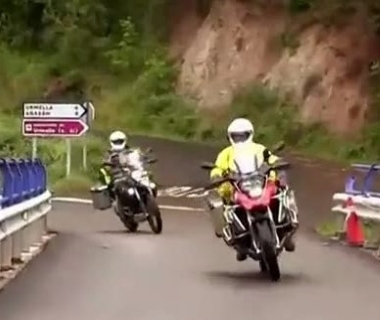}
&
\includegraphics[width=0.22\textwidth]{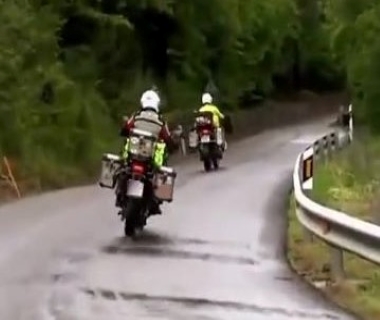}
\end{tabular}
\caption{Representative frames of two temporal sequences from the OVIS dataset where AirObject fails to identify the objects.}
\label{fig:limitation}
\end{figure}

\subsection{Limitations}

Although our proposed method, AirObject, achieves the state-of-the-art performance for object identification across challenging scenarios, its success is partially based on the segmentation head's success. Furthermore, we notice that it struggles to identify objects when two distinct parts of an object are observed across the temporal sequences. For example, in \fref{fig:limitation}, we show representative frames of two temporal sequences. Here, intuitively one would match the bikes across the two temporal sequences using either the color of the rider's coat or the relative spatial position. However, since AirObject only uses a self-attention limited to the object, it fails to identify the bikes. We believe this is a limitation of our approach. In this context, an interesting avenue for future work would be to explore encoding temporal relative spatial information using cross-attention across adjacent objects within the scene and the background.

\section{Conclusion}

Discriminatively identifying objects is a critical and challenging problem for robotics tasks involving autonomous exploration and semantic localization \& mapping. In this paper, we present a novel temporal object encoding method, AirObject, to generate global object descriptors. The proposed temporal encoding method accumulates structural knowledge across evolving topological graph representations of an object. Our experiments show that the proposed method leads to state-of-the-art performance in object identification on four challenging datasets. We also showcase that our AirObject descriptors are robust to severe occlusion, viewpoint shift, deformation, and scale transform. While we present the efficacy of AirObject for object identification, we envision AirObject to play a key role in endowing robots with general class-agnostic semantic knowledge for real-world robotic applications.

\section*{Acknowledgements}

\noindent This work was supported by ONR Grant N0014-19-1-2266 and ARL DCIST CRA award W911NF-17-2-0181.

{\small
\bibliographystyle{ieee_fullname}
\bibliography{egbib}
}

\end{document}